\documentclass{article}
\usepackage{template, hyperref, amsfonts, amsmath, graphicx, lineno, xcolor, setspace}

\title{Stochastic Latent Transformer: Efficient Modelling of Stochastically Forced Zonal Jets}

\author{
Ira J. S. Shokar$^1$, Rich R. Kerswell$^1$, Peter H. Haynes$^1$  \\
$^1$Department of Applied Mathematics and Theoretical Physics, University of Cambridge,\\ Wilberforce Road, Cambridge, CB3 0WA, UK\\
}

\begin{document}
\begin{spacing}{1.15}
\maketitle

\begin{abstract}
We present a novel probabilistic deep learning approach, the 'Stochastic Latent Transformer' (SLT), designed for the efficient reduced-order modelling of stochastic partial differential equations. Stochastically driven flow models are pertinent to a diverse range of natural phenomena, including jets on giant planets, ocean circulation, and the variability of midlatitude weather. However, much of the recent progress in deep learning has predominantly focused on deterministic systems. The SLT comprises a stochastically-forced transformer paired with a translation-equivariant autoencoder, trained towards the Continuous Ranked Probability Score. We showcase its effectiveness by applying it to a well-researched zonal jet system, where the interaction between stochastically forced eddies and the zonal mean flow results in a rich low-frequency variability. The SLT accurately reproduces system dynamics across various integration periods, validated through quantitative diagnostics that include spectral properties and the rate of transitions between distinct states. The SLT achieves a five-order-of-magnitude speedup in emulating the zonally-averaged flow compared to direct numerical simulations. This acceleration facilitates the cost-effective generation of large ensembles, enabling the exploration of statistical questions concerning the probabilities of spontaneous transition events.
\end{abstract}

\section{Introduction}

The computational expense of modelling turbulence has spurred research into more computationally cost-effective alternatives known as reduced-order models (ROMs) to advance simulations of fluids \cite{DOF_turbulence}. In this context, deep learning methods, particularly autoencoders \cite{deeplearning, VAE}, have surpassed traditional techniques such as Proper Orthogonal Decomposition (POD) \cite{POD_turbulence} in constructing spatial ROMs given to their capacity for nonlinear transformations \cite{Linot, intrinsic_dynamics}. Deep learning has also been employed to model the temporal evolution of turbulent fluid flows, significantly reducing computational costs during inference \cite{fluids_ML_Ricardo_Brunton, reinforcementturb, osti_1852843, osti_1924060, yousif_zhang_yu_vinuesa_lim_2023, deepmindnow} - with efforts combining these methods with autoencoders to evolve dynamics in latent space \cite{Linot, vaetransformer2023, effective_dynamics, physfluidstrans} outperforming linear methods such as dynamic-mode decomposition (DMD) \cite{DMD}. The aforementioned studies have primarily focused on deterministic nonlinear partial differential equations (PDEs), however in many physical applications, including fluid mechanics, state-dependent stochastic terms are often included to represent unresolved processes such as small-scale turbulent eddies.

\begin{figure}[t!] 
    \includegraphics[width=\textwidth]{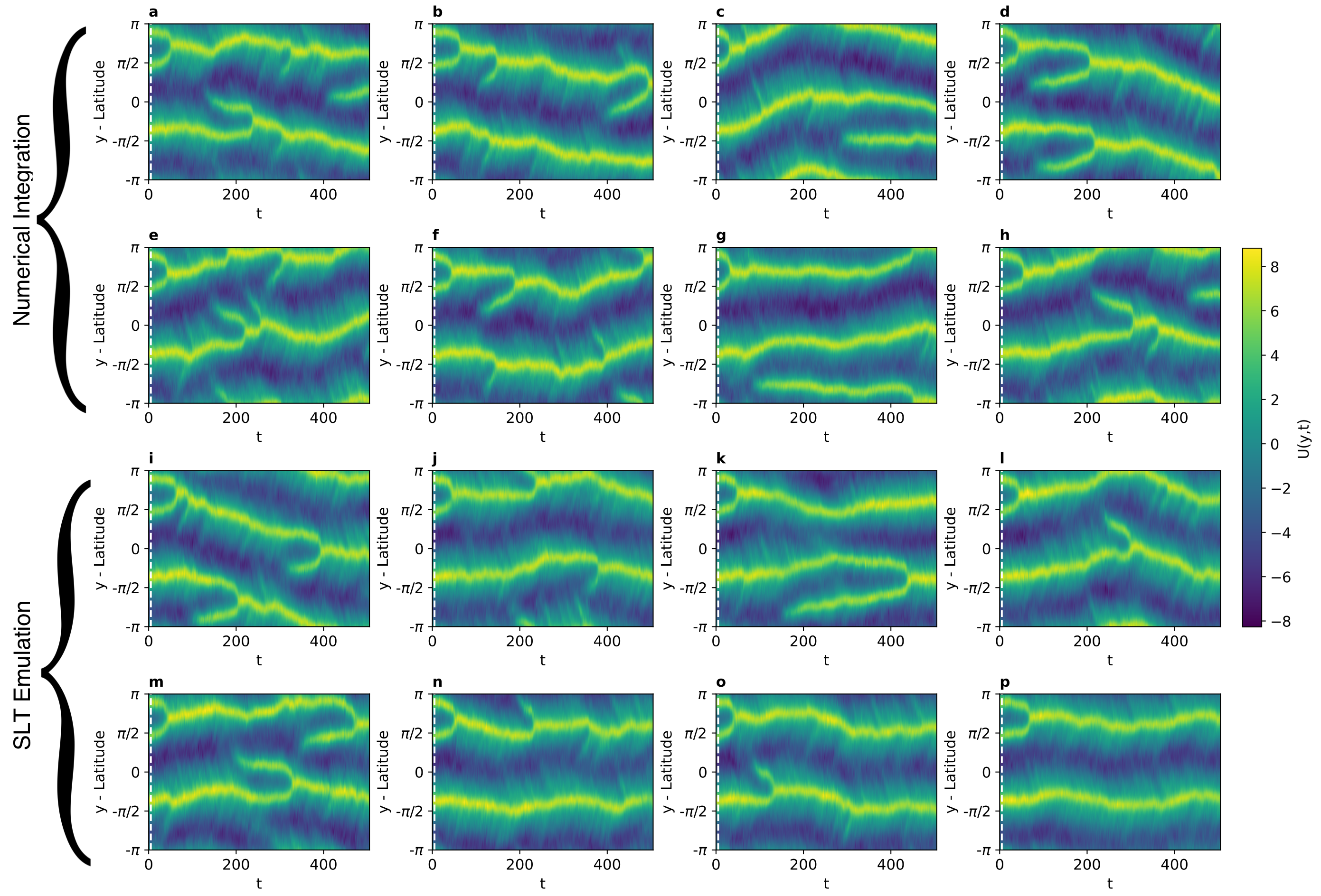}
    \caption{\small\textbf{Latitude-time plots of $\mathbf{U(y,t)}$ displaying ensemble of numerical integrations and neural network emulations.} \textbf{a-h} exhibit numerical integrations with identical initial conditions up to $t=10$ (indicated by the dotted line) and distinct realisations of random noise, $\xi$, after $t=10$, spanning a forecast period of 500 time units. \textbf{i-p} showcase neural network emulations with identical initial conditions as (\textbf{a-h}), but with different noise histories $\epsilon \sim \mathcal{N}(0,1)$ after $t=10$. Numerical integration employs a time step of $\delta t=4\times10^{-4}$, with \textbf{a-h} displaying coarsened time intervals, with the neural network operating at unit time intervals $\Delta t=1$. Yellow indicates positive $U(y,t)$, signifying eastward jets. Noteworthy events within these jets encompass nucleation, coalescence, and latitudinal translation. The neural network adeptly captures anticipated system features, offering plausible evolutions.}
    \label{fig:num_int_and_ml_emul}
\end{figure}

In this study, we develop a deep learning approach tailored to handle Stochastic Partial Differential Equations (SPDEs). We use a translation-equivariant autoencoder to find a reduced order representation and employ a stochastically forced transformer to model the latent dynamics in a probabilistic manner. While Operator Learning methods \cite{FNO} have excelled in modelling PDEs \cite{hu2022neural, operators_pde, operatorexteremes, FourCastNet}, for tackling SPDEs we focus on transformer-based methods \cite{Attention} due to their proficiency in discerning correlations within an input set using scaled dot-product attention. This allows the model to generate a context-aware weighted sum of its inputs when forecasting the next sequence member \cite{Attention}. In the context of modelling SPDEs this attends to previous time histories of the flow as well as forcing noise to produce a state-dependent weighting of the forcing contribution.

One domain in which there is demand for ROMs of stochastically-driven flows is weather and climate modelling, characterised by turbulent and complex multi-scale flows. Parameterisation schemes, often explicitly stochastic, are employed to approximate the effects of important small-scale processes not explicitly resolved by the spatial resolution of the model, such as cloud physics, radiative transfer, and turbulent eddies \cite{Berner2017}, with stochasticity here accounting for the chaotic behaviour of the unresolved dynamics. Recent studies have leveraged deep learning methods to create stochastic parameterisations for coarsened deterministic models \cite{Zanna, perezhogin2023generative, Gagne_2020}. These introduce stochasticity on the basis that no deterministic function of resolved variables can fully account for the effects of unresolved variables. In contrast, our study focuses on employing deep learning to emulate systems that explicitly incorporate a stochastic component, that drives the system dynamics.

In this work, we investigate producing a deep learning ROM of a well-studied idealised Geophysical Fluid Dynamics (GFD) system - beta-plane turbulence \cite{rhines_1975, zonal_jets}. This system serves as a useful simplest model for many aspects of flows in atmospheres and oceans. This is a single-layer flow in which turbulence is generated via stochastic forcing, $\xi(x, y, t)$, in the vorticity equation:

\begin {align}
\partial_t \zeta+ \textbf{u} \cdot \nabla \zeta + \beta \partial_x \psi = \xi - \mu \zeta + \nu_n \nabla^{2n} \zeta,
\end{align}

with the velocity field defined as $\mathbf{u}(x, y, t) = (-\partial_y \psi, \partial_x \psi)$ (note the sign convention used widely in GFD), where $\psi$ is the stream function and the relative vorticity is $\zeta = \partial_x v - \partial_y u$. The parameter $\beta$ represents the variation of rotation with latitude, while energy dissipation occurs through linear damping at rate $\mu$ at large scales, while hyperviscosity, with coefficient $\nu$ and order $n$, is used to remove vorticity at small scales. 

In turbulence, a Reynolds decomposition is typically employed as an averaging procedure to separate coherent structures from fluctuations. Here we apply an eddy-mean decomposition, $u(x,y,t) = \overline{u}(y,t) + u'(x,y,t)$, \cite{beta_turb, beta_bouchet} in the zonal direction to separate the dynamics, giving equations for the zonally averaged zonal velocity field $U(y, t)=\overline{u}(y, t)=\frac{1}{L_x}\int^{L_x}_{0}\textbf{u}dx$ (where $L_x=2\pi$ is the domain size in $x$), and the associated vorticity fluctuation fields, $\zeta^{'}(x, y, t)$:

\begin{align}
\partial_t U&=-\mu U + \nu_n \partial_y^{2 n} U + \overline{\zeta^{'} v^{'}}\\
\partial_t \zeta^{'}&=-U \partial_x \zeta^{'}+\left(\partial_{y y} U-\beta\right) \partial_x \psi^{'}+\xi-\mu \zeta^{'}+ F_e + \nu_n \nabla^{2 n} \zeta^{'}
\end{align}

where $F_e= \partial_y\left(\overline{\zeta^{'} v^{'}}\right)-\partial_y\left(\zeta^{'} v^{'}\right) -\partial_x\left(\zeta^{'} u^{'}\right)$ are nonlinear terms. Here we see that the evolution depends on the nonlinear two-way interaction between the mean flow and the eddies and that the applied noise, $\xi$ is applied only directly applied the eddy fields, and therefore is filtered in the mean flow.

This study focuses on a particular choice of parameters, $\beta=90$, $\mu=4\times10^{-2}$, $\nu=100$, and an energy injection rate of $\varepsilon=10^{-4}$, with a principal forcing wavenumber of $k_f=16$ and a time step of $\delta t=4\times10^{-2}$, chosen to be in the regime in which coherent zonal jets form. The parameter choice made here leads to the jets exhibiting a wide range of time variability behaviour \cite{Sukioransky}. The latitude-time plots in Figure \ref{fig:num_int_and_ml_emul}.a-h, generated via numerical integration, illustrate how the system self-organises to produce jets - the inclusion of $\beta$ disrupts the turbulent energy cascade, leading to jet formation (defining a jet as a local maxima of $U$). Each plot displays numerical integrations with identical histories up to the dotted line, with the subsequent dynamics driven by different realisations of $\xi$. These jets exhibit complex phenomena, including spontaneous transition events such as nucleation (the formation of a new jet), coalescence (the merging of two jets) and latitudinal drifting of persistent zonal jets. In this regime it may be argued that the model provides an analogue of week-to-week variations in the large-scale dynamics of the tropospheric jet, a key driver of midlatitude weather variations \cite{vallis}.  Details of the particular numerical scheme used are given in the Appendix.

In looking to model the mean flow using deep learning without explicitly parameterising equation (3), we combine equations (2) and (3):

\begin{align}
\partial_t U(y, t)= \left(-\mu + \nu_n \partial_y^{2 n}\right)U(y, t) + \int_T \mathcal{G}[U, \zeta'(\xi); \beta, \mu, \nu_n] dt,
\end{align}

where $\mathcal{G}$ is a non-linear operator that represents the 
two-way interaction between the eddies and the the large-scale flow to provide the evolution of $U(\mathbf{y}, t)$. 

Existing research on deep learning applied to stochastic parameterisation
focuses on acquiring a stochastic specification solely for the eddy effects at any instant in terms of the large-scale flow \cite{Zanna, perezhogin2023generative}. For the system stated here this would equate to a learning a data-driven form of the second term on the right-hand side of equation (4) offline, before using this closure to solve equation (4) using a PDE solver. By contrast, we propose a novel deep learning approach designed to learn the equivalent to the entire right-hand side of equation (4) - a stochastic specification for how the SPDE governing $U$ should be updated in time. We observe that the evolution of the system, $\partial_t U$, depends on the temporal histories of $U$, $\zeta'$, and $\xi$. Consequently, our deep learning model is structured to take short-term histories of $U$ and incorporates an attention mechanism that implicitly learns an effective form of the interaction between these temporal histories and the mean flow, here represented by $\mathcal{G}$, which enables the prediction of possible realisations of $\partial_t U$.

We demonstrate that this deep learning approach is capable of reproducing both short-term and long-term statistics of the original system. Illustrating a faithful model over very-long time evolutions has not been previously demonstrated using transformers for PDEs or SPDEs, and this could not be replicated for this system by other state-of-the-art generative models, such as Variational Autoencoders (VAE) \cite{VAE} and adversarial models such as those used in \cite{Gagne_2020, perezhogin2023generative}. The deep learning approach provides a cost-effective means to generate large ensembles, enabling us to study regimes close to spontaneous transition events, such as nucleation and coalescence events. This allows us to approach previously computationally prohibitive questions, such as quantifying to what extent different events are driven by deterministic or stochastic dynamics.

\section{Methods}

Recent advances in fluid modelling using transformers have been constrained by the large number of degrees of freedom of these spatiotemporal systems. In weather system nowcasting \cite{FourCastNet, ClimaX, Pangu-Weather}, researchers have omitted using attention to model temporal correlations and focus on spatial features by adapting the vision transformer \cite{ViT}. In our approach, we maintain the ability to address both spatial and temporal correlations, necessary for this system. We achieve this by first obtaining a reduced-dimensional representation of the original system \cite{Linot, intrinsic_dynamics, effective_dynamics, embeddings} as described by equation (4), which is input to a transformer to evolve the stochastic latent dynamics.

\begin{figure}[t!] 
    \begin{center}
        \includegraphics[width=\textwidth]{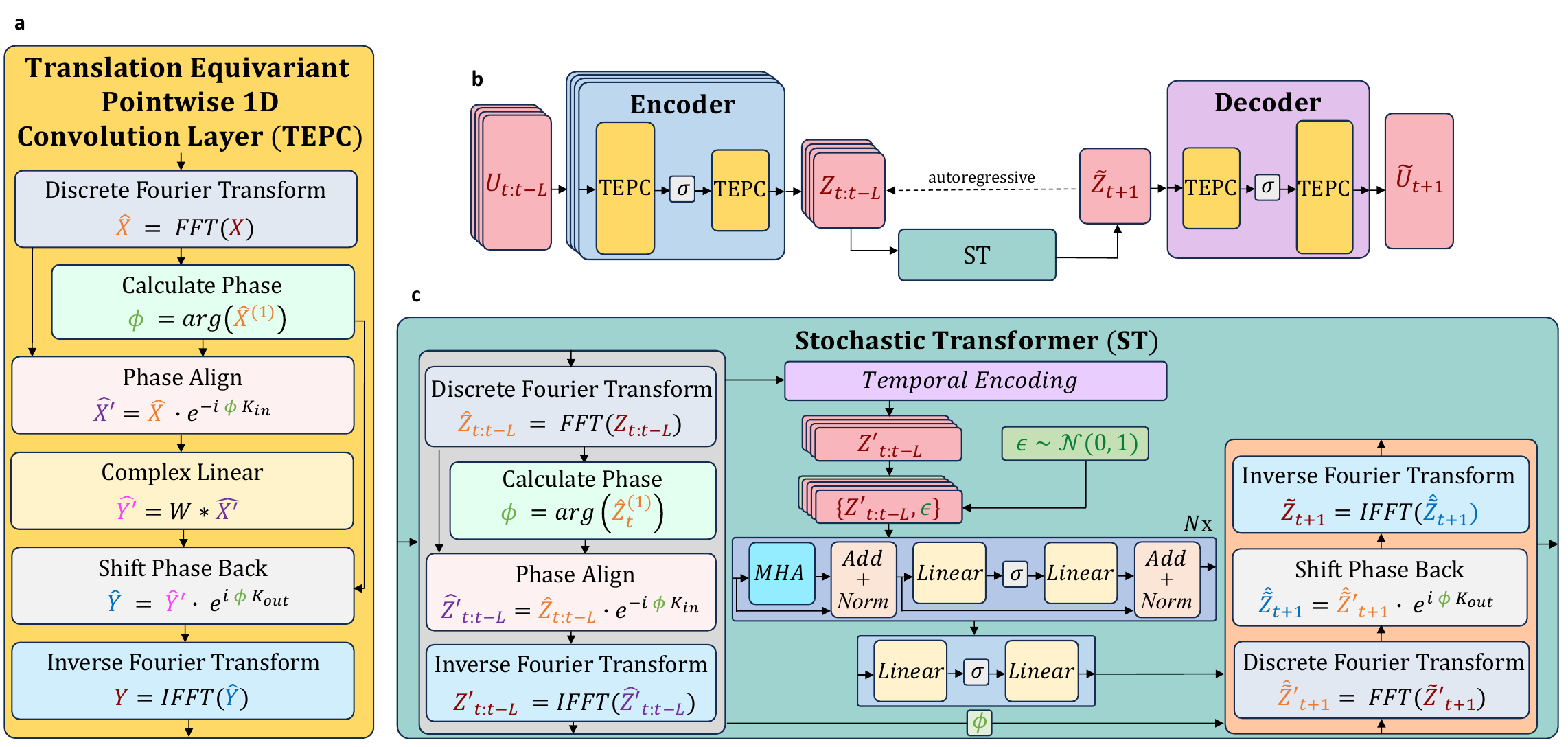}
    \end{center}
    \caption{\small \textbf{Schematic of the Stochastic Latent Transformer (SLT) architecture and its components. a} Translation Equivariant Pointwise 1D Convolution (TEPC) Layer. The layer convolves inputs $X$ with learned weights $W$ in the frequency domain, with weights $W$ learned irrespective of the phase, $\phi$ ($\phi$ is the argument of the first mode of $\hat{X}$). \textbf{b} Stochastic Latent Transformer (SLT) architecture. Solid arrows indicate the forward pass. The encoder employs two TEPC layers with a nonlinear activation function, $\sigma$, to encode the short time history of $U_{t-L:t}$. The resulting $Z_{t-L:t}$ is fed into the Stochastic Transformer (ST) for forecasting $\tilde{Z}_{t+1}$. The dotted line indicates the autoregressive flow ($\tilde{Z}_{t-L+1:t+1}$) for forecasting subsequent steps up to $\tilde{Z}_{t_{max}}$ during inference. The decoder then transforms $\tilde{Z}_{t+1}$ (or $\tilde{Z}_{t+1:t_{max}}$) back to the physical space $\tilde{U}_{t+1}$ (or $\tilde{U}_{t+1:t_{max}}$), mirroring the encoder's architecture. \textbf{c} Stochastic Transformer (ST) architecture. The weights are translation invariant, as with the TEPC, with the phase, $\phi$, at time $t$ removed. The green block represents the random noise vector, $\epsilon \sim \mathcal{N}(0, 1)$, and is concatenated with the latent space-time histories $Z_{t-L:t}$. The architecture consists of N transformer blocks, where 'MHA' is multi-headed attention using scaled dot-product attention (see Figure \ref{fig:mha} for details), 'Linear' is learned linear transformation and $\sigma$ is the nonlinear activation function. Further details are outlined in the Methods section.}
    \label{fig:ml_arch}
\end{figure}

\subsection{Autoencoder - Translation Equivariant Pointwise Convolution}

Given the multi-scale nature of fluid flows, Convolutional Neural Networks (CNNs) \cite{cnn} are commonly used to capture spatial correlations. Each convolutional layer preserves translational equivariance through circular padding. However, it is important to note that while individual convolution operations exhibit translational equivariance, CNNs as an architecture lack global translational equivariance due to max pooling and striding operations used for dimensionality reduction \cite{cnn_translation}. 

Assuming spatial homogeneity for the forcing term $\xi$, the beta-plane system demonstrates latitudinal symmetry under the transformation $(t, x, y, \psi, \zeta) \mapsto (t, x, -y, -\psi, -\zeta)$. Translation equivariance in the beta-plane system can be denoted by $\phi U(y, t) = U(\phi y, t)$, where $\phi$ signifies a shift operator in the $y$ dimension, a result of the system's periodic boundary conditions and inherent symmetries. Given that this is an important property of the system, maintaining translational equivariance is advantageous (conversely, in cases with different boundary conditions, where this symmetry may not exist, the network should adapt accordingly and not exhibit such equivariance).

To address this, we introduce a phase-equivariant convolutional architecture for both the encoder and decoder, consisting of what we shall refer to as Translation Equivariant Pointwise 1D Convolution Layers (TEPC), see Figure \ref{fig:ml_arch}.a for illustration. Consequently, the resulting latent space, $Z \in \mathbb{R}^{D_{\mathcal{M}}}$, remains equivariant to any shift $\Delta y$ in the physical space. We draw inspiration from both the method of slices \cite{mos_1, mos_2} by learning translation invariant weights, employed by \cite{Linot}, as well as the Fourier Neural Operator \cite{FNO} by performing a convolution as a linear transformation in Fourier space.

The method of slices involves taking a discrete Fourier transform, $\mathcal{F}$, to transform the input data, $X$, at each TEPC layer into the spectral domain, $\hat{X} = \mathcal{F}(X) \in \mathbb{C}$. We obtain the phase information, $\phi$, from the first Fourier mode using $\phi = \text{arg}(\hat{X}^{(1)})$. We store this phase and construct a phase-aligned solution, $\hat{X}' = \hat{X}e^{-\phi \cdot K_y}$, such that the first Fourier mode is a pure cosine. This step ensures that the weights $\textbf{W}$ are calculated to be invariant to this phase. Subsequently, we apply the convolution operation in the form of a linear transformation in Fourier space, $\hat{H}' = \textbf{W} \ast \hat{X}'$. We reintroduce the phase information via $\hat{H} = \hat{H}'e^{-\phi \cdot K_{out}}$ before transforming the result back to the physical space, $H = \mathcal{F}^{-1}(\hat{H}) \in \mathbb{R}$.

We denote the TEPC layer operation as $H = \Gamma(X, \textbf{W})$. The encoder and decoder each consist of a two-layer structure, with a nonlinear activation function, $\sigma = \text{GELU}$ \cite{gelu}, following the first layer. We can express the encoder as $Z = \mathcal{E}_{\phi}(U) = \Gamma_2[\sigma \Gamma_1(U, \textbf{W}_1), \textbf{W}_2] \in \mathbb{R}^{D_{\mathcal{M}}}$, where $\phi$ represents the weights $\textbf{W}_1 and \textbf{W}_2$. The decoder, $U = \mathcal{D}_{\theta}(Z)$, with weights $\theta$, takes the same form but to transform from latent space back to physical space, $\mathbb{R}^{D_{\mathcal{M}}} \rightarrow \mathbb{R}^{256}$. The point-wise convolution allows for both local and global convolutions to take place in a single procedure, allowing the TEPC to reduce input dimensions to any desired size, in contrast to CNNs which rely on pooling operations to achieve a specific reduction factor.

Through experimentation, it was observed that the TEPC only required two layers with a single activation function to establish mappings between physical and latent spaces, as illustrated in Figure \ref{fig:ml_arch}.b, resulting in fewer operations and faster inference compared to a CNN with the same latent dimension reduction. 

The latent space, $Z$, maintains translational equivariance with respect to the input $U$. Alternatively, not reintroducing the phase $\phi$ in the final layer achieves a translation-invariant latent representation of the input field. This approach, when extended to handle 2D and 3D inputs, holds great promise for computer vision tasks where properties such as translation invariance and equivariance are desirable, an example being object recognition. Additionally, it eliminated the need for data augmentation during training, thereby reducing our overall training costs. 

\subsection{Stochastic Latent Transformer}
 
The encoder provides a latent representation of $U \rightarrow Z$. In order to evolve the system forward in time we look to model the stochastically forced dynamical system $Z(t+\tau) = F[Z(t), \epsilon]$ on $\mathcal{M}$. One motivation for using transformers is that they have been shown to outperform Recurrent Neural Networks (RNN) \cite{RNN} by effectively handling temporal dependencies in sequences by utilising the scaled dot-product attention mechanism that considers the entire sequence and has achieved improved training speed over RNNs by parallelising backpropagation through multi-headed attention which evaluates multiple attention operations in parallel.

Figure \ref{fig:ml_arch}.c illustrates the transformer architecture that comprises a multi-head attention block and a feed-forward neural network. The attention blocks allow the model to weight the importance of different parts of the input sequence when making predictions, while the feed-forward network (comprising two linear layers with a non-linear activation function, here $\sigma = \text{GELU}$, between the two) allows it to learn complex non-linear relationships. The transformer blocks and feed-forward network are each concatenated with skip connections around them to enhance gradient propagation towards the input during the backpropagation phase of training.

We employ a regression transformer, adapting the original transformer architecture proposed in \cite{Attention}, by exclusively utilising the transformer decoder, which throughout this article, we shall simply refer to as the transformer. This choice stems from the fact that our input data is in the form of a time series of continuous values, rather than a prompt that necessitates semantic comprehension, as is in natural language processing. The original transformer was used for the classification of discrete data in natural language tasks, and so we adapt the final layers by replacing the softmax function in the feed-forward network with two linear layers with a non-linear activation function, again $\sigma = \text{GELU}$, between the two to provide a regression transformer. 

\begin{figure}[t!] 
    \begin{center}
        \includegraphics[width=\textwidth]{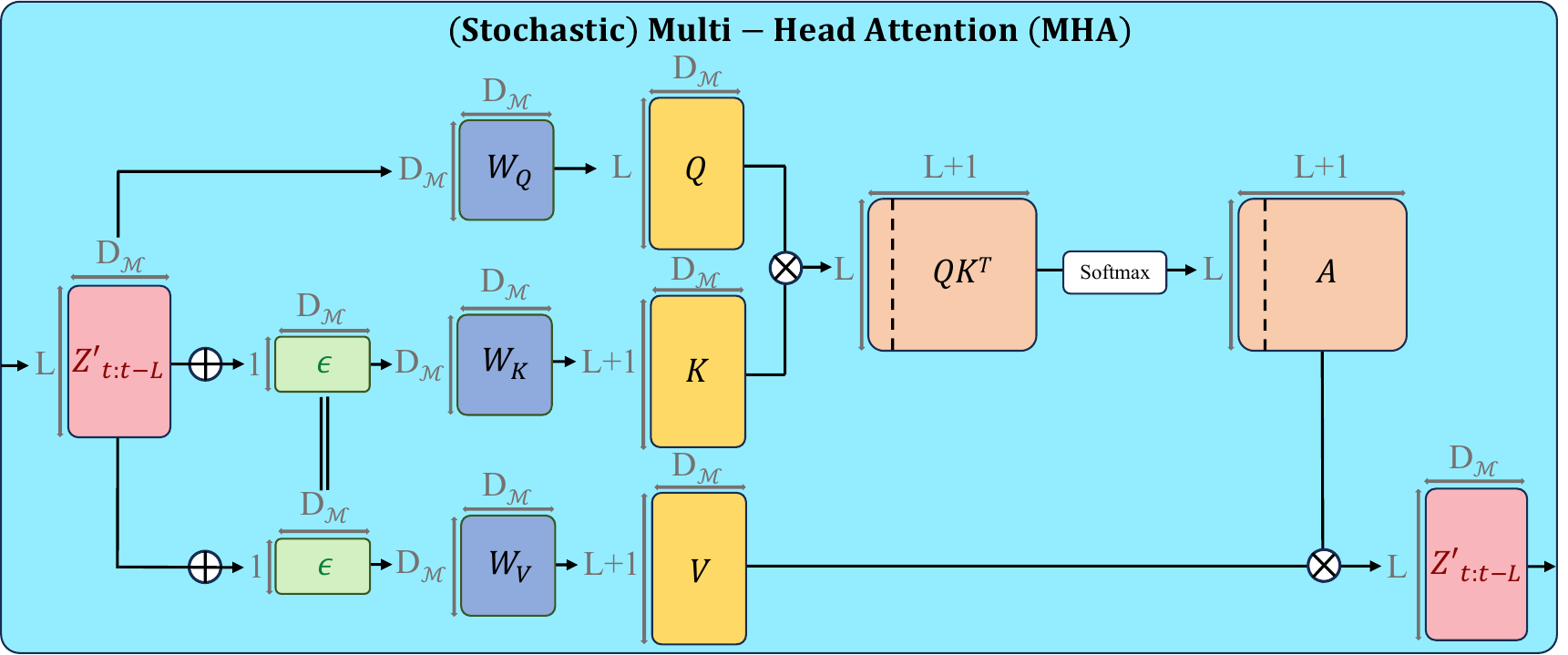}
    \end{center}
    \caption{\small \textbf{Schematic of stochastic multi-headed attention.} The initial transformer block among the N blocks incorporates a stochastic variant of multi-headed attention. Here $\epsilon \sim \mathcal{N}(0, 1)$ is introduced as an additional sequence member, $L$ represents the length of the latent time history, and $D_{\mathcal{M}}$ denotes the latent dimension. Weights, W, linearly transform inputs into Q (Query), K (Key), and V (Value) vectors. The $\epsilon$ input is identical for both the K and V vectors. Dotted lines in the A (Attention) matrices separate the additional row introduced from cross-attention, given forcing $\epsilon$, and the self-attention of the latent time histories. This illustrates the case where the batch size = 1 and number of heads = 1, with reshaping implemented to extend this to multiple heads. The non-stochastic self-attention in the following transformer blocks operates in the exact same manner, simply without the concatenation of $\epsilon$.}
    \label{fig:mha}
\end{figure}

In the natural language setting, the softmax layer gains probabilities for the next possible token, sampled in inference to make a prediction such that the most probable word is not always selected to produce diverse outputs. By removing the softmax and sampling operations, we must now induce stochasticity in a different manner. Current efforts in stochastic transformers for regression tasks explore different approaches: using stochastic activation functions \cite{lcutransformer}, replacing attention with stochastic alternatives for computational efficiency \cite{sinkformers}, or introducing noise to all inputs \cite{MCTransformer}. In contrast, our approach incorporates forcing noise as an additional sequence element, $\{Z_{t:t-L}, \epsilon\}$, such that we only impose a single forcing when making a forecast of $Z_{t+1}$. The attention mechanism in the transformer identifies correlations among the input elements, using this to calculate the state-dependent weighting of the forcing noise $\epsilon \sim \mathcal{N}(0, 1)$ when predicting $Z_{t+1}$ based on the latent flow history $Z_{t:t-L}$. This facilitates ensemble generation with different noise realisations, akin to sampling $\xi$ in equation (4).

We employ two different attention mechanisms: self-attention and cross-attention. While the computation of the attention layer remains consistent across both cases, the inputs to the layer differ - see Figure \ref{fig:mha} for schematic of the cross-attention implemented. The attention operation can be defined as:

\begin{align}  
\text{Attention}(Q, K, V) = A(Q, K)V =\text{softmax}\left(\frac{QK^T}{\sqrt{d_k}}\right)V 
\end{align}

Here, Q (query vector), K (key vector), and V (value vector) are obtained through linear transformations of their respective inputs: \(Q = MW_Q\), \(K = NW_K\), and \(V = PW_V\). \(d_k\) represents the dimension of the key vector, K, and \(\text{softmax}\) denotes the row-wise softmax function. The application of softmax introduces sparsity to the attention matrix, $A$, regulating the emphasis placed on each input in the sequence of length $L$. A notable characteristic of the attention mechanism is that the vectors are calculated dynamically based on the correlating patterns within the key (K) and query (Q) vectors.

In self-attention, the inputs $M$, $N$, and $P$ share the same values, which in this context \(M = N = P =Z_{t:t-L}\). In cross-attention, $N$ and $P$ take the same values, while $M$ differs. Our aim for the attention mechanism to appropriately weight the significance of input features, which here take the form of the latent mean flow \(Z_{t:t-L}\) and the forcing noise \(\epsilon \sim \mathcal{N}(0, 1)\). As such, we choose that $M = Z_{t:t-L}$, while $N = P = \{Z_{t:t-L}, \epsilon\}$. Within cross-attention, we effectively conduct self-attention on \(Z_{t:t-L}\), capturing temporal correlations, while providing an adaptive weighting of the stochastic forcing, \(\epsilon\), contingent on the time history \(Z_{t:t-L}\). This can be conceptualised as determining the degree of emphasis to allocate on \(\epsilon\) in relation to the current system state. Analogously, this mirrors phase space regimes in the beta-plane system, with transitions between statically stable fixed points induced by fluctuations, relative to the current state in phase space \cite{trans_freddy}.

It's worth noting that only the initial multi-headed attention layer in the transformer incorporates cross-attention with the forcing term \(\epsilon\). Subsequent layers utilise self-attention, to again ensure that we only impose a single forcing when making a forecast of $Z_{t+1}$, to match how the beta-plane system is forced, as described in equation (1).

We add a time-embedding vector \cite{timevec} to each input latent-space vector to incorporate temporal order information into the model, with the exception of the $\epsilon$ to indicate that $\epsilon$ is not part of the temporal sequence. During inference, the transformer evolves autoregressively in the latent space to produce $Z^{(i)}_{t+1:t_{max}}$ for each ensemble member $i$, before using the Decoder to produce the evolution $U^{(i)}_{t+1:t_{max}}$, see Figure \ref{fig:ml_arch}.b for illustration. We denote the stochastic transformer operation $Z_{t+1} = \mathcal{T}_\varphi[Z_{t:t-L}, \epsilon]$, parameterised by weights, $\varphi$. 

As with the autoencoder, we wish for the transformer to also act in a translation-invariant manner. The transformer architecture itself is not translation invariant due to fully connected layers within the architecture. To overcome this we only input to the transformer phase aligned solutions, using the method of slices and then add back the phase to each output. This way the transformer handles inputs that have been translated in an identical manner (given identical samples of noise $\epsilon$). Here we separate each value of $Z$ into a phase-aligned solution, $Z'$ and phase, $\phi$. The phase is extracted from the latent time history $Z_{t:t-L}$ from the sequence member at the time $t$ only, $\phi(t)$. This is because removing the phase from the previous time steps in the time history would prevent the transformer from also learning the latitudinal drifting dynamics that we observe in the system. Each sequence member is then phase-aligned with respect to $\phi(t)$, $Z_{t:t-L}' = Z_{t:t-L}e^{-\phi(t) \cdot K_z}$. We reintroduce the phase information to the output from the transformer $Z_{t+1}'$ via $\phi(t)$, $Z_{t+1} = Z_{t+1}'e^{\phi(t) \cdot K_z}$.

With the transformer evolving the system forward in an autoregressive fashion during inference, we must apply this each time step forward, as the transformer expects to see inputs where $Z_t$ has zero-phase in the first Fourier mode at every time step. We must phase align solutions with forecasting each autoregressive time step $Z_t+n$ before input to the transformer, to ensure that data given to the transformer during inference is from the distribution of the training data.

\subsection{Training}

In most research on data-driven modelling of fluid flows, the Mean Squared Error (MSE) has been the predominant choice as the loss function. This is because it aligns with the goal of minimising the discrepancy between predicted outputs and the reference data. However, MSE proves inadequate for handling probabilistic forecasts, which are essential due to the unpredictable and random nature of turbulence, especially in the context of the beta-plane turbulence model.

As a training objective, we employ the Continuous Ranked Probability Score (CRPS) \cite{CRPS}, a proper scoring rule and a commonly used metric for probabilistic forecasts in meteorology. \cite{probforecastgenscore} demonstrated the use of the CRPS as an alternative to adversarial training methods - these employ competitive learning but can suffer from issues such as non-convergence and mode collapse \cite{GAN}. The CRPS is a generalisation of the Mean Absolute Error (MAE) for distributional forecasts, and can be expressed as:

\begin{align}
\text{CRPS}(F, \tilde{U})&=-\int_{-\infty}^{\infty}(F(\tilde{U})-1_{\{\tilde{U} \geq U\}})^2 d\tilde{U} \\
&=\mathbb{E}[|\tilde{U}-U|]-\frac{1}{2} \mathbb{E}\left[\left|\tilde{U}-\tilde{U}^{\prime}\right|\right]
\end{align}

We use the non-parametric estimate to the CRPS \cite{fair}, estimating F with the empirical cumulative distribution function (CDF) of $m$ independent and identically distributed (iid) samples $\tilde{U} \sim F$:

\begin{equation}
\mathrm{CRPS}(U, \tilde{U})=\underbrace{\frac{1}{m} \sum_{i=1}^m\left| U-\tilde{U}^{(i)}\right|}_{\text{MAE}} - \underbrace{\frac{1}{2m^2} \sum_{i=1}^m \sum_{j=1}^m\left|\tilde{U}^{(i)}-\tilde{U}^{(j)}\right|}_{\text{Ensemble Variation}} 
\end{equation}

where $U$ is the truth trajectory from the training dataset, $\tilde{U}^{(i)}$ is the $i^{th}$ member of the prediction ensemble, where the model produces an ensemble of size $m$ for time step $t+1$. The network is trained to minimise the difference between each predicted ensemble member and the truth trajectory $U_{t+1}$, while simultaneously maximising the dissimilarity between each individual ensemble member $\tilde{U}^{(i)}_{t+1}$ due to the inclusion of the Ensemble Variation term in equation (8). This promotes output diversity, overcoming mode collapse issues seen in adversarial training. 

To ensure that the Autoencoder approximates the identity function, that is map from physical space to the latent space and back, without evolving the system forward in time, we introduce an additional term $\mathrm{MAE}(U_t,\hat{U}_t) = \left|U_t-\hat{U}_t\right|$ where $\hat{U}_t=\mathcal{D_{\theta}}[\mathcal{E_{\phi}}(U_t)]$ is the output of the Autoencoder without being passed through the transformer. Importantly, the encoder, decoder and transformer are trained simultaneously, to ensure that the encoder and decoder learn temporal features when finding the latent representation, however, this additional loss constraint ensures that the encoder and decoder are only performing a spatial transformation. 

While research conducted by \cite{FNO} found that nonlinear activation functions could recover energy at higher wavenumbers, following a filtering operation, our attempts to replicate these findings were unsuccessful. Consequently, the TEPC layers in our model do not function as filters for higher modes; instead, they serve to reduce the number of modes to a latent dimension. Generative deep learning models have been shown to emphasise larger-scale features due to the diminished contribution of smaller-scale features, inadequately capturing higher frequency modes \cite{specbias}. In the context of modelling physical data, the preservation of not only large-scale properties but also energy across all scales is essential. To address the recovery of higher wavenumbers, we introduce a term into the loss function, serving as a soft constraint, ensuring the conservation of the absolute value of each Fourier mode. This involves employing the discrete Fourier transform, denoted as $\mathcal{F}$, to transform $U_t$ and $\hat{U}_t$ and subsequently taking the modulus, resulting in $|\mathcal{F}[U_{t}]| \in \mathbb{R}$ and $|\mathcal{F}[\hat{U}_{t}]| \in \mathbb{R}$. Taking the $\mathrm{MAE}(|\mathcal{F}[U_{t}]|, |\mathcal{F}[\hat{U}_{t}]|) = \left||\mathcal{F}[U_{t}]|-|\mathcal{F}[\hat{U}_{t}]|\right|$ derives a spectral loss term, ensuring that the autoencoder preserves energy at all necessary scales.

Non-linear transformations are not distance-preserving, however, this is an advantageous property to have, both for interpretability of the latent space, but we also observed that it aids in numerical stability during training. We assert that two states in physical space, if close together, should also be close together in our latent space. This is not hard constrained in the model, instead, we add an additional term to the loss function that intends to minimise the distance in latent space between our prediction, in latent space, $\tilde{Z}_{t+1}$, and the encoded representation of $Z_{t+1}:=\mathcal{E}_{\phi}({U}_{t+1})$ from our 'truth' training set, while retaining output diversity - it is, therefore, appropriate to use again use the CRPS in latent space. This leads to the following objective function:

\begin{equation}
\mathcal{L} = \mathrm{CRPS}\left(U_{t+1}, \tilde{U}_{t+1}\right) + \mathrm{CRPS}\left(Z_{t+1}, \tilde{Z}_{t+1}\right) + \mathrm{MAE}\left(U_{t}, \hat{U}_{t}\right) + \mathrm{MAE}\left(|\mathcal{F}[U_{t}]|, |\mathcal{F}[\hat{U}_{t}]|\right)
\end{equation}

This loss function is applied over each batch, and the gradient with respect to model parameters is calculated to perform backpropagation updates. Through experimentation, we found there was no requirement to weight these terms to achieve convergence during training. 

We use $2\times\text{10}^{\text{6}}$ unit time intervals for training and 2,000 unit time intervals for validation to monitor overfitting. The network receives randomly sampled mini-batches as inputs, which are short-time evolution histories of length $L$ of $U(y, t:t-L)$ separated by unit time intervals of the zonally averaged zonal velocity from numerical integration, as defined in equation (4). While a time step of $\delta t=4\times10^{-2}$ is required for stability of the numerical integration, sampling at such small intervals is unnecessary for the neural network to capture the dynamics. Therefore data input to the network is sampled at unit time intervals (2500 times larger than the numerical integration time step). Each time step of the system is represented as a one-dimensional vector $U(y, t) \in \mathbb{R}^{256}$, with the target output being the subsequent unit time step $U(y, t+1)$, also obtained from the numerical integration.

The numerical integrations were conducted using the GeophysicalFlows \cite{GFF} package in Julia \cite{Julia}. Machine Learning models were built using the PyTorch \cite{torch} framework and all analyses were conducted in Python \cite{python}.
\subsection{Model Hyperparameters}

We ran a hyperparameter sweep We ran a Bayesian hyperparameter sweep \cite{wandb} to determine the best-performing model. The metric used to minimise was the Helliger Distance, defined in section 3.2, between PDFs, produced via long-time evolutions of $20,000$ from a numerical integration and the output from the deep learning model. Through this sweep we found that the following hyperparameters were optimal: batch size: 128; optimiser: Adam; initial transformer learning rate: 5e-4; initial autoencoder learning rate: 2.5e-3, learning rate scheduler: exponential decay with decay rate: $\gamma$=0.9825; training epochs: 400, convolution filters in each TEPC layer: 4, CRPS ensemble size: 4; $L$ (length of time history used to make next forecast): 10, number of transformer blocks: 3.

\section{Results}

\subsection{Short Term Evaluation}

Given the system's stochastic nature, we compare an ensemble of model forecasts to an ensemble of numerical integrations, each with a different realisation of forced noise, $\xi$, (shown in Figure \ref{fig:num_int_and_ml_emul}.a-h) to evaluate the model's ability to learn the evolution of $U(y, t)$, as opposed to a direct comparison between individual trajectories. The bottom 8 plots (i-p) in Figure \ref{fig:num_int_and_ml_emul} depict 8 emulations from the SLT, with the same time histories as (a-h) input to the model and subsequent dynamics driven by different realisations of $\epsilon$ for 500 time units. The SLT yields plausible evolutions when compared to the numerical integrations, effectively predicting the imminent coalescence event and capturing subsequent nucleating events, additional coalescence events, and latitudinal translation in a realistic manner.

To make a more quantitative assessment of the deep learning model's performance, we employ metrics for short and long forecast periods. For short forecast periods, the trajectory distribution is narrow, rendering the CRPS as a suitable probabilistic measure. We compute the CRPS between the emulated ensemble (Figure \ref{fig:num_int_and_ml_emul}.i-p) and a single truth trajectory, specifically the numerical integration in Figure \ref{fig:num_int_and_ml_emul}.a. While a lower CRPS indicates better performance, it is most useful to compare the CRPS of the ensemble forecast to that of ensemble numerical integrations, both with respect to the same reference trajectory. We separate the CRPS, equation (8), into its MAE and ensemble variation components for visualisation in Figure \ref{fig:evals_long}.

\begin{figure}[t!] 
    \includegraphics[width=\textwidth]{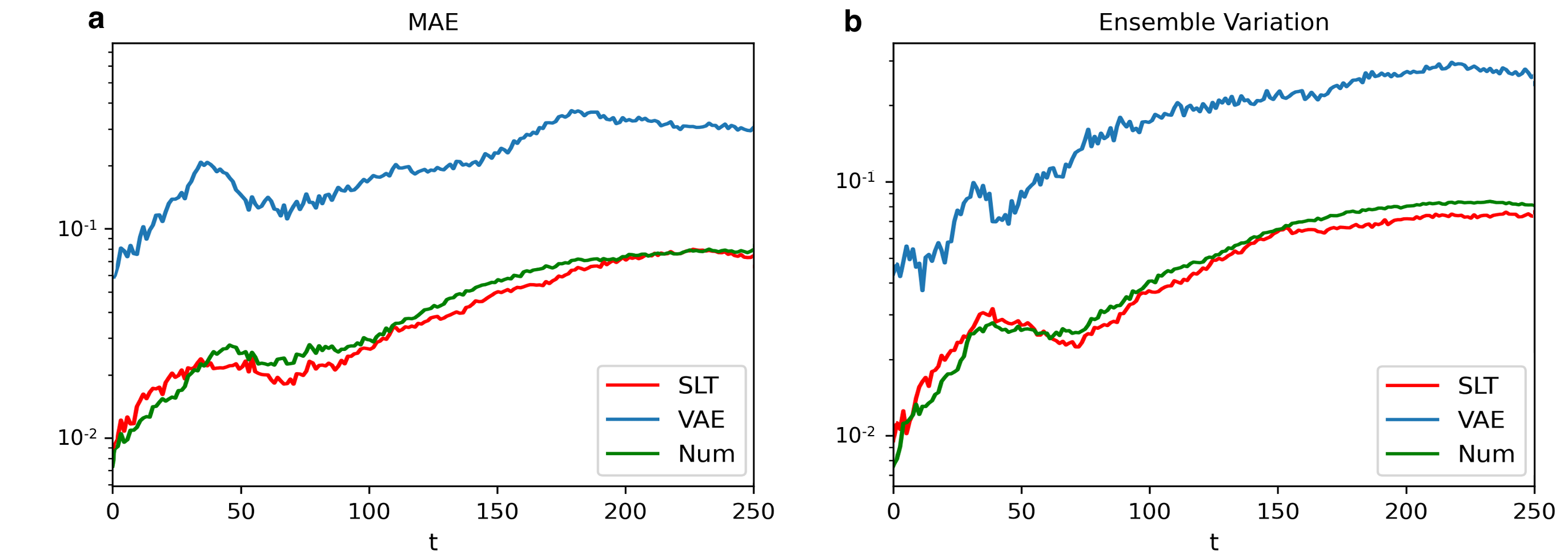}
    \caption{\small\textbf{Performance evaluation over short time scales. a} presents the Mean Absolute Error (MAE) for short-term tracking ability. Green represents an ensemble of 7 numerical integrations, while red shows 7 emulations from the SLT with respect to the reference truth trajectory (Figure \ref{fig:num_int_and_ml_emul}.a). Blue indicates an ensemble produced by a Variational Autoencoder (VAE). \textbf{b} showcases ensemble variation. Using these evaluation metrics, we observe that the SLT exhibits strong agreement with the numerical integration.}
    \label{fig:evals_short}
\end{figure}

Figure \ref{fig:evals_short}.a displays the MAE over time, in red for the SLT ensemble and in green for an ensemble of the 7 other numerical integrations (Figures \ref{fig:num_int_and_ml_emul}.b-h). The SLT exhibits strong agreement with the numerical integration throughout the period. Both reach saturation at $\sim t$=250, indicating that trajectories have sufficiently diverged and MAE is no longer a meaningful metric. We also compare the outputs of a temporal Variational Autoencoder (VAE) \cite{VAE}, a popular generative model architecture, shown in blue. The VAE demonstrates notably poorer performance, as evidenced by the larger MAE values. Performing the same analysis with the ensemble variation term, displayed in Figure \ref{fig:evals_short}.b. The SLT again shows good agreement with the numerical integration, while the VAE shows a much greater forecast spread. In the Supplementary Information, we demonstrate the SLT's similar effectiveness for trajectories with different initial conditions.

One goal of this research was to create a computationally more efficient model than numerical integration. Each emulation in Figure \ref{fig:num_int_and_ml_emul}.i-p took 8.35 milliseconds to produce 500 future time units using the SLT, while each numerical integration in Figure \ref{fig:num_int_and_ml_emul}.a-g took 29.8 minutes to produce a time series over the same period: the SLT being $\sim\text{2.13x10}^{\text{5}}$ times faster on identical hardware (NVIDIA P100) running CUDA 10.2.

\subsection{Long Time Evolutions}

\begin{figure}[t!] 
    \includegraphics[width=\textwidth]{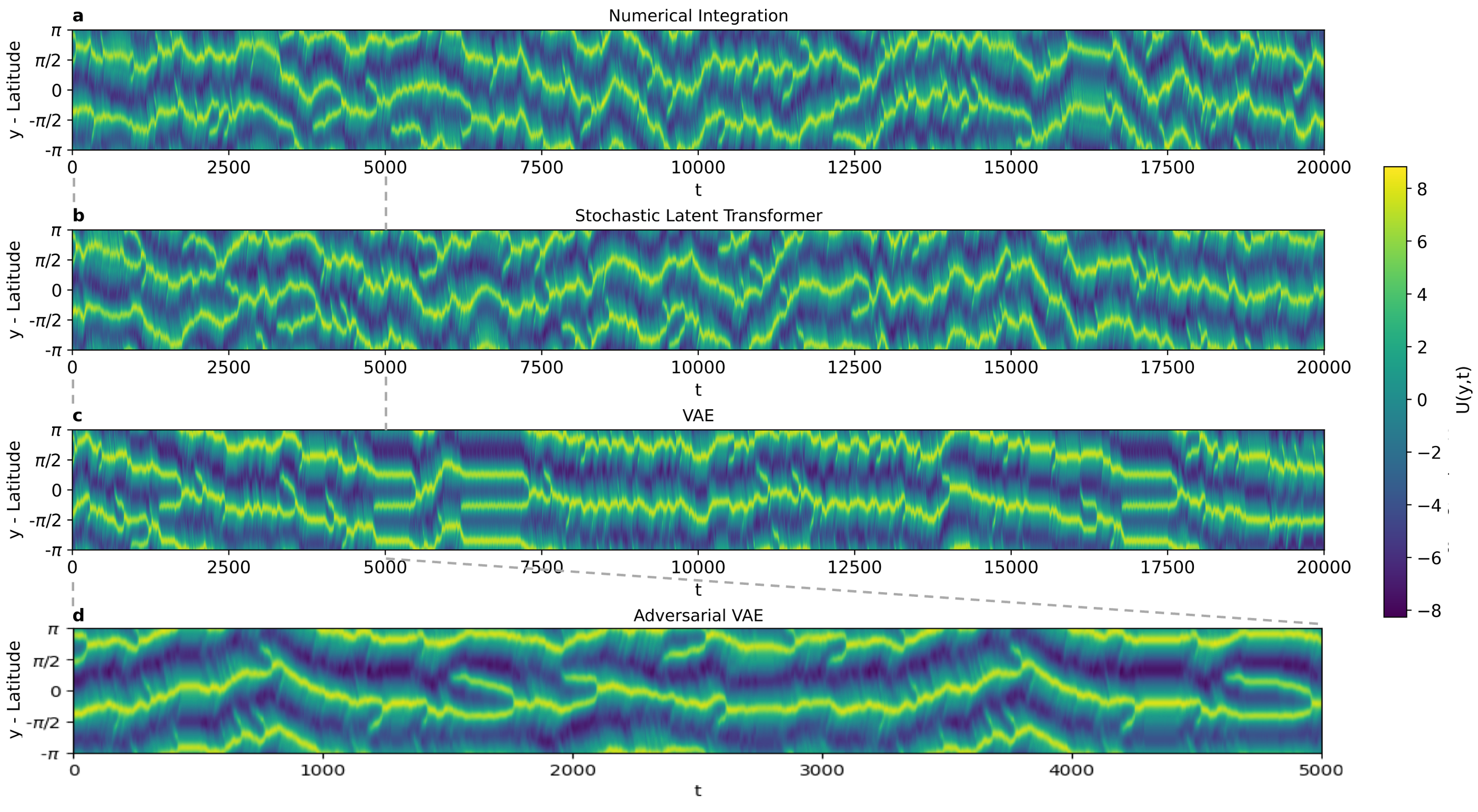}
        \caption{\small\textbf{Assessing model faithfulness over long-term evolutions.} \textbf{a} presents a long-term evolution obtained through numerical integration, spanning 20,000 observable time units (from a longer evolution of $t_{\text{max}}=10^\text{6}$). \textbf{b} displays a long-term evolution generated by the SLT, showing qualitative similarity to \textbf{a} with the occurrence of nucleation and coalescence events, as well as latitudinal translation. \textbf{c} illustrates a long-term evolution produced by a Temporal Variational Autoencoder (VAE) that becomes unstable. \textbf{d} depicts a long-term evolution spanning 5,000 time increments generated by a VAE with adversarial training, which suffers from mode collapse, resulting in cyclically repeating features. These emulations demonstrate the robustness of the autoregressive SLT in capturing and maintaining realistic long-term dynamics, which could not be replicated by existing architectures.}
    \label{fig:long_emulations}
\end{figure}

To assess the SLT's performance over extended evolutions, we conduct an emulation for 20,000 time units, shown in Figure \ref{fig:long_emulations}.b, and compare with a numerical integration spanning the same duration, shown in Figure \ref{fig:long_emulations}.a. Given the rapid divergence of trajectories under different noise realisations, initial conditions become less significant and an ensemble is not required, with the long evolutions capturing the broad spectrum of system dynamics. The results from the SLT are show good agreement with the numerical integration with Figure \ref{fig:long_emulations}.b, demonstrating consistent and expected dynamics throughout the integration period.

This performance, however, could not be replicated with alternative generative models, such as the temporal Variational Autoencoder (VAE) in Figure \ref{fig:long_emulations}.c. The VAE exhibits nonphysical behaviour at $ \sim t$=5000 while displaying a lower frequency of nucleation, coalescence events, and latitudinal translation compared to numerical integration. These discrepancies stem from significant deviations between VAE outputs and the training data distribution, as evident from the short-term errors in Figures \ref{fig:evals_short}.a-b. Although an adversarial VAE in Figure \ref{fig:long_emulations}.d produces qualitatively plausible results, it experiences mode collapse, resulting in cyclically repeating features over a shorter evolution period. These limitations are not observed with the SLT architecture, trained using the CRPS. Further comparisons between the SLT architecture and these alternative models are discussed in Appendix B.

For a quantitative assessment of longer-term evolutions, we examine statistical properties of the flow. We generate Probability Density Functions (PDFs) for individual $U$ values and their derivatives, $\partial_y U$ and $\partial_t U$. These PDFs serve to measure the learned spatial and temporal correlation. We conduct this analysis over a time integration period of $200,000$ time units, producing two PDFs: $p(U)$ representing the PDF from numerical integration, $U$, and $q(\tilde{U})$ representing the PDF of forecasted values derived from the SLT, $\tilde{U}$. 

\begin{figure}[t!] 
    \includegraphics[width=\textwidth]{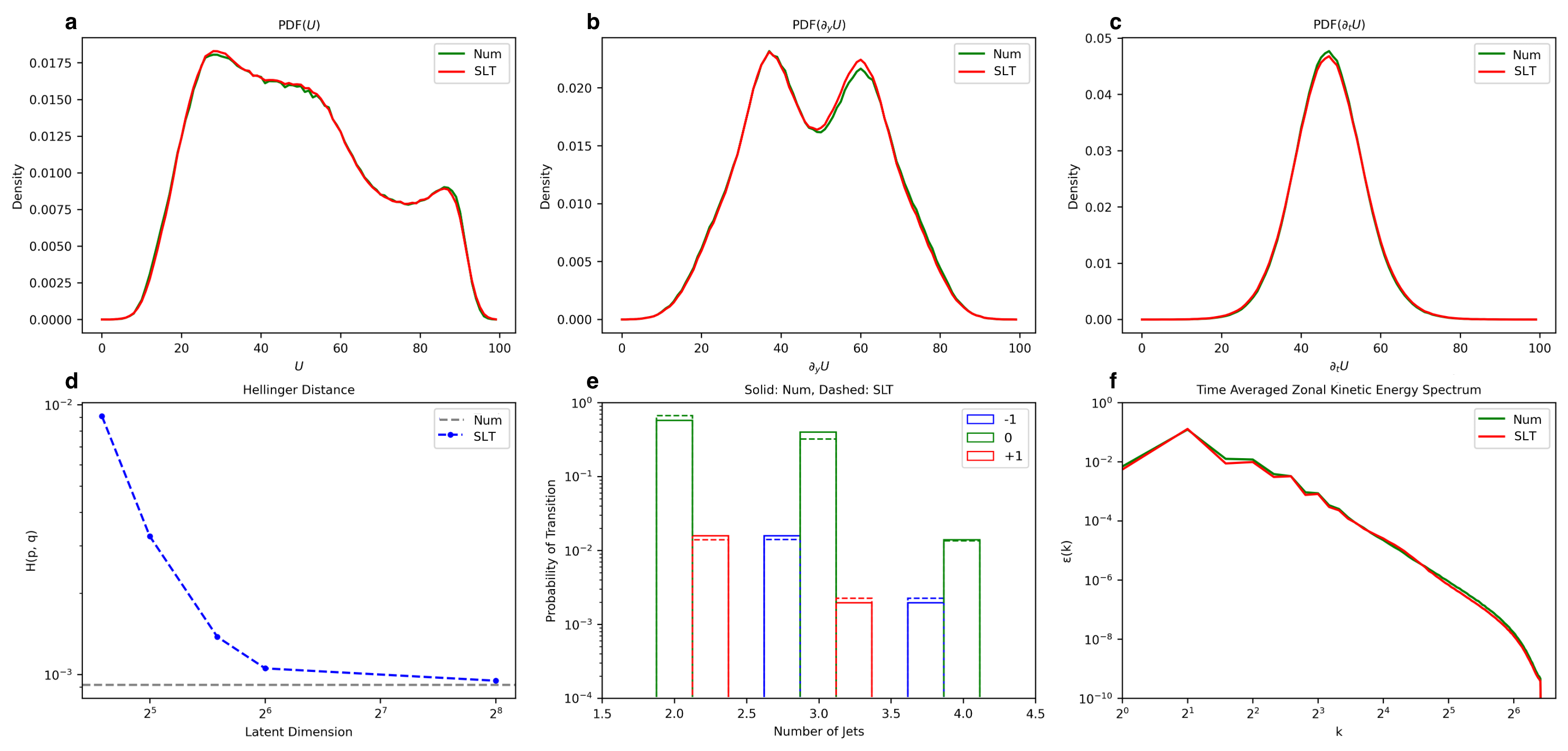}
        \caption{\small\textbf{Performance evaluation of long time evolutions. a} PDF of values of $U$, from the numerical integration in green and the SLT in red, obtained from long evolutions of $\text{10}^\text{6}$ time units. \textbf{b} same as \textbf{a} for values of $\partial_y U$, to demonstrate learned spatial correlations. \textbf{c} same as \textbf{a} and \textbf{b}, for values of $\partial_t U$, to demonstrate learned temporal correlations. \textbf{d} illustrates the Hellinger distance $H(p, q)$ between PDFs of $U$, $\partial_y U$, and $\partial_t U$ obtained from long evolutions of $\text{10}^\text{6}$ time units from the SLT, as the latent dimension increases (blue). It also shows $H$ between two numerical integrations with different noise realisations for comparison (grey). \textbf{e} depicts the PDF for the likelihood of the system transitioning in the number of jets, based on $\text{10}^\text{6}$ time units observable time from both numerical integration (solid lines) and SLT (dashed lines). The PDFs are derived from jet count frequencies at observable time $t$, showing whether the count increases (red), decreases (blue), or remains constant (green) at observable time $t+1$. \textbf{f} time averaged power spectral density displayed for both the numerical integration and the SLT, showing the energy content of each wavenumber in the zonally-averaged zonal direction, $U$. Using these evaluation metrics, we observe that the SLT again exhibits strong agreement with the numerical integration over long time evolutions. }
    \label{fig:evals_long}
\end{figure}

In Figures \ref{fig:evals_long}.a-c, we present the distributions of individual constituents of $p(U)$ and $q(\tilde{U})$: $U$, $\partial_y U$, and $\partial_t U$, along with their corresponding $\tilde{U}$ equivalents. When assessing the disparity between distributions, the comparison of distribution tails serves as a valuable metric to determine the emulator's efficacy in representing the entire distribution and accurately capturing rare event statistics. Figures \ref{fig:evals_long}.a-c show good agreement between numerical integration (in green) and SLT (in red), showcasing an appropriate representation of distribution tails. This suggests that the SLT successfully captures events with limited representation in the training dataset, demonstrating its capacity to faithfully approximate the entire distribution. In particular, our observations indicate that the transformer has learned the exact form of the temporal correlations exact form, showcasing minimal discrepancy between the PDFs in Figure \ref{fig:evals_long}.c. Figure \ref{fig:si_pdf}, in the Supplementary Figures, displays the joint PDFs between each of these components.

We quantify the distance between these distributions, following \cite{Gagne_2020}, using the Helliger Distance, $H$, as our measure:

\[ H(p, q) = \frac{1}{2} \int \left\|p(\textbf{U}) -q(\mathbf{\tilde{U}}) \right\|d\textbf{U}. \]

where $textbf{U} = {U, \partial_y U, \partial_t U}$. In Figure \ref{fig:evals_long}.d, we observe how $H$ changes with different sizes of the latent dimension $D_\mathcal{M}$, with a distinct change in the gradient ($dH/dD_{\mathcal{M}}$) when $D_\mathcal{M} = 64$. For all evaluation results presented in this paper, we set $D_\mathcal{M} = 64$, as only marginal improvements in $H$ are observed when increasing the degrees of freedom. This corresponds to a compression factor of 1024, when compared to the numerical integration, as the numerical integration is required to integrate over a 2D domain, $\mathbb{R}^{256 \times 256} \rightarrow \mathbb{R}^{D_{\mathcal{M}} = 64}$. Figure \ref{fig:evals_long}.c also depicts the $H$ value between two trajectories, generated through numerical integration (indicated by the grey dashed line), giving context to the values of $H$ obtained. Using $H$ as a metric, we can directly compare outputs across various models and hyperparameter selections, as detailed in the section 2.4.

We can measure the duration spent in each jet configuration and the occurrence frequency of state transitions (coalescence or nucleation events). PDFs of the number of jets at time $t$ and changes in jet count at time $t+1$ are generated from longer evolutions spanning $\text{10}^\text{6}$ time units, both from numerical integration and the SLT. In Figure \ref{fig:evals_long}.e the frequencies of these PDFs closely align between the numerical integration (solid lines) and the SLT (dashed lines), indicating that the SLT adequately reproduces the underlying system. 

Neural networks have been shown to exhibit a spectral bias \cite{specbias}, wherein they tend to excel in capturing low-frequency modes but struggle to represent higher frequencies that contribute less to the overall energy. The spectral loss term, outlined in section 2.3, is employed to combat this by penalising outputs from the decoder that exhibit spectral deviation from the training data. In Figure \ref{fig:evals_long}.f, the time-averaged power spectral density is plotted for both the numerical integrations (in green) and the autoregressive outputs of the SLT (in red). We observe a high level of agreement across all scales, paying particular attention on the higher wavenumbers where data-driven approaches often fail. These findings, in conjunction with the aforementioned evaluation metrics, allow one to place confidence in the SLT's capacity to faithfully emulate the dynamics of the system under consideration. This positions the SLT as a valuable tool for investigating dynamical phenomena observed in the beta-plane system, facilitated by its computational speed-up.

\subsection{Characterising Transition Events}

\begin{figure}[t!] 
    \includegraphics[width=\textwidth]{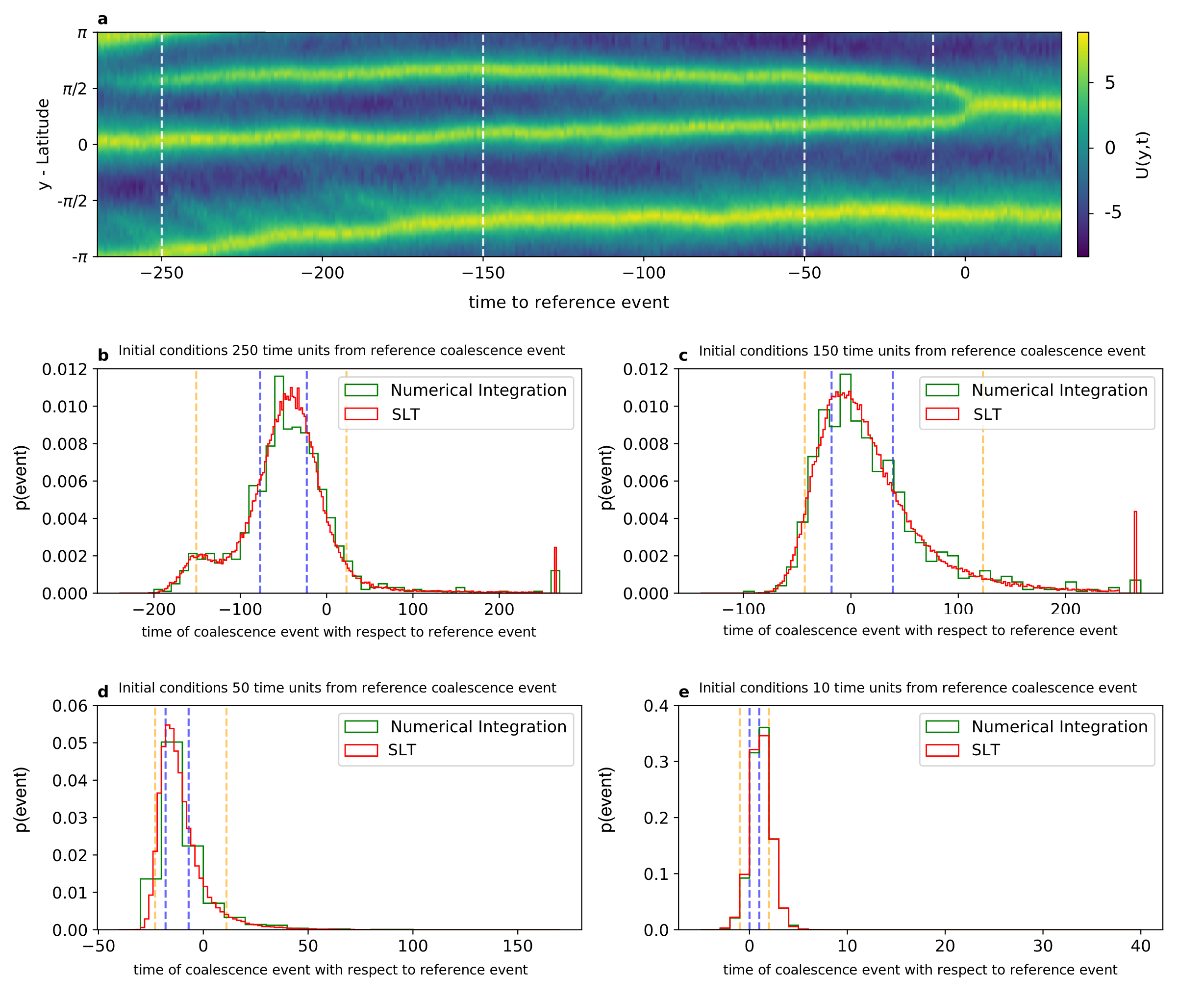}
    \caption{\small\textbf{PDFs of time to coalescence events.} \textbf{a} displays a reference coalescence event generated through numerical integration, with dashed lines representing initial conditions located 250, 150, 50, and 10 observable time units away from the event. Note that the x-axis is shifted with 0 at the occurrence of the coalescence event. \textbf{b-e} depict PDFs of the time taken between each initial condition and the coalescence event. The ensemble consists of $\text{10}^\text{6}$ instances generated using the SLT (red) and $\text{10}^\text{3}$ instances obtained through numerical integration (green). The orange dashed lines indicate the $\text{5}^{th}$ and $\text{95}^{th}$ quartiles of the probability density, and we display the corresponding time within this bound. The blue lines represent the $\text{25}^{th}$ and $\text{75}^{th}$ quartiles for the red curve. As the initial conditions approach the reference event, the PDFs become sharper, and the dynamics become increasingly deterministic. The green and red curves show agreement, indicating that the SLT successfully captures the expected distributions.}
    \label{fig:pdfs_merge}
\end{figure}

In chaotic or stochastically driven systems, predicting transition events, such as nucleation and coalescence in the beta-plane system, is of interest. To accurately determine lead times from initial conditions, ensembles are required to account for the system variability, however, if the numerical solvers are too computationally expensive, generating large ensembles for such studies might not be feasible. Previous work has been conducted to artificially upsample the frequency of these spontaneous transition events using a rare-event algorithm \cite{Bouchet_rare} or augmenting the training data for deep learning of extremes \cite{extremenudge}. The introduction of the SLT now allows us to efficiently generate large ensembles of trajectories without the prohibitive costs associated with numerical solvers.

\begin{figure}[t!] 
\includegraphics[width=\textwidth]{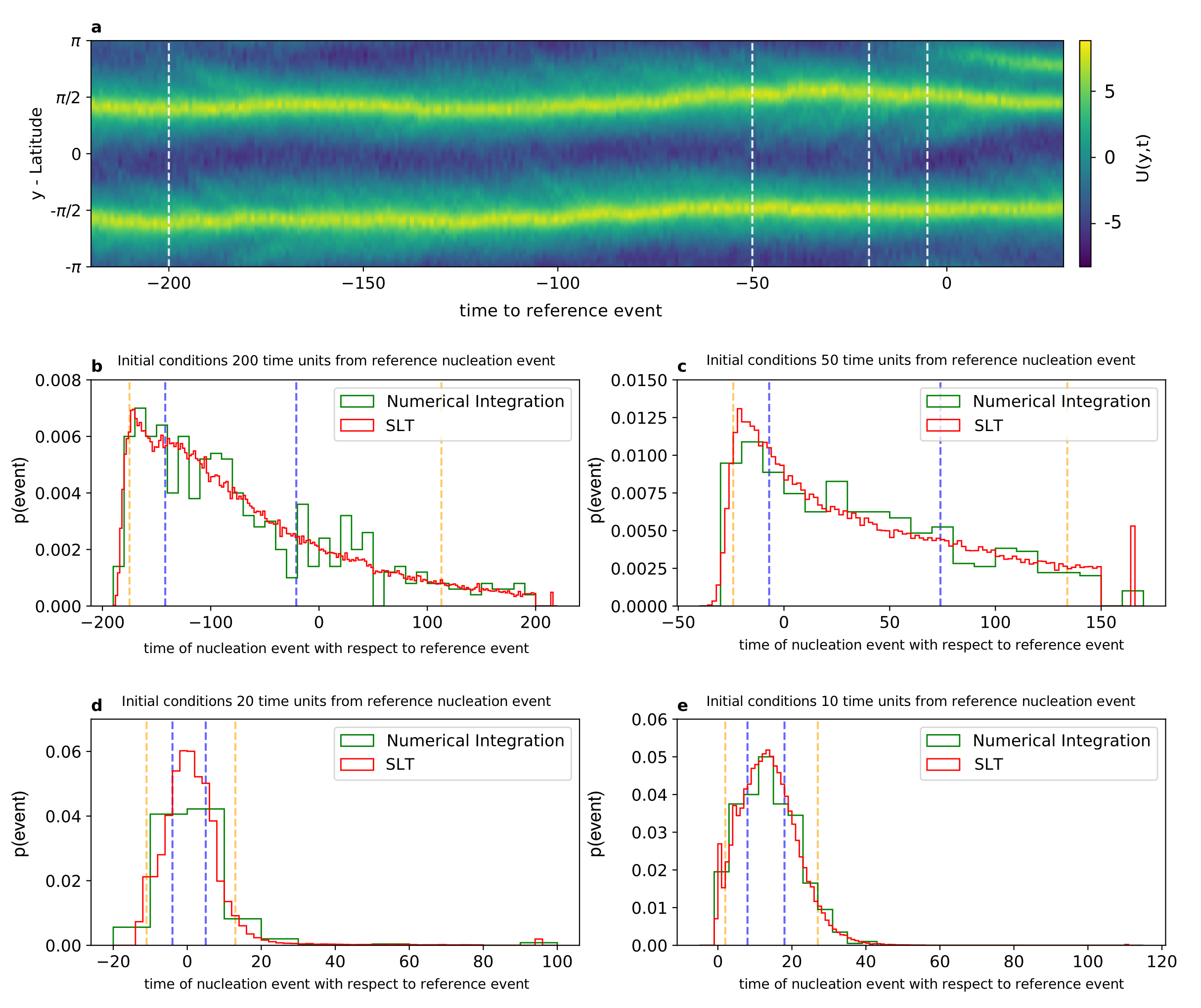}
\caption{\small\textbf{PDFs of time to a nucleation event.} Here we show the same information as in Figure \ref{fig:pdfs_merge}, but for a reference nucleation event, shown in \textbf{a}, with the dashed lines referencing a number of initial conditions that are $250, 150, 50$ and $10$ observable time units from the coalescence event. While we can see that as the initial conditions are moved closer to the reference event the PDFs in \textbf{b-e} become much sharper, the time scales over which this takes place indicate that the nucleation events are driven much more by stochastic dynamics than the coalescence events. Again, we also see agreement between the green and red curves indicating that the SLT is successfully capturing the expected distributions.}
\label{fig:pdfs_nuc}
\end{figure}

In this experiment, we use two reference events generated by numerical integration: a coalescence event shown in Figure \ref{fig:pdfs_merge}.a and a nucleation event shown in Figure \ref{fig:pdfs_nuc}.a. For the coalescence event, we analyse various initial conditions set at different lead times prior to this event, at 250, 150, 50, and 10 time units before the reference event, indicated by the dashed lines in Figure \ref{fig:pdfs_merge}.a. To estimate the probability density function (PDF) representing the time interval between each initial condition and a coalescence event for each trajectory, we perform forward integrations, each using an ensemble of $\text{10}^\text{6}$ trajectories with the SLT. The resulting PDFs are displayed in Figures \ref{fig:pdfs_merge}.b-e.

As the initial conditions approach the reference coalescence event, the Probability Density Function (PDF) progressively narrows, signalling a shift towards more deterministic behaviour in the system. This transition can be quantified by examining the variance of each PDF distribution. For instance, with initial conditions at $t=150$ units from the reference event as shown in Figure \ref{fig:pdfs_merge}.c, the SLT provides a 90\% probability of a coalescence occurring within the time range of 110-273 units, as denoted by the span between the orange dashed lines in Figures \ref{fig:pdfs_merge}.b-e. As the initial conditions move closer to the event, this time window tightens to 21-53 units when $t=50$ units away, and further to 9-13 units when $t=10$ units from the event. In Figures \ref{fig:pdfs_merge}.b-e, the blue lines signify a 50\% probability of occurrence, with the width of this interval also contracting progressively as the initial conditions approach the reference event.

To validate the PDFs produced by the SLT, we generate ensembles of size $\text{10}^\text{3}$ through numerical integration, represented by the green plots in Figure \ref{fig:pdfs_merge}. The close agreement observed between the PDFs affirms the reliability of the results and underscores the SLT's effectiveness as an emulator for predicting transition events. Generating each ensemble of size $\text{10}^\text{6}$ for 500 time units in Figure \ref{fig:pdfs_merge}.b using the SLT only required 16 minutes on a two GPUs (NVIDIA P100). In contrast, producing the $\text{10}^\text{3}$ ensemble through numerical integration took around 10 days on identical hardware, making this analysis impossible for very large ensemble sizes.

Figure \ref{fig:pdfs_nuc}.a displays the reference nucleation event and by analysing the PDFs in Figures \ref{fig:pdfs_nuc}.b-e, we observe a considerably broader dispersion and a distinct skew when the reference event is positioned 200 and 50 time units away, in contrast to the coalescence events in Figure \ref{fig:pdfs_merge}. Irrespective of the lead time to the reference event, there is a notable probability of a nucleation event occurring shortly after the initial conditions. This implies that initial conditions with a 2-jet configuration exist in a phase space region more sensitive to stochastic forcing and thus more susceptible to undergo a transition than a 3-jet state.

Comparing the breadth of each PDF for nucleation and coalescence events reveals notable distinctions. When considering initial conditions at $t=50$ units from the reference nucleation event (see Figure \ref{fig:pdfs_nuc}.c), the time span corresponding to a 90\% probability of occurrence extends to 254 forecasted time units, contrasting with a span of only 33 time units for the coalescence event (see Figure \ref{fig:pdfs_merge}.d) When only $t=10$ away from their respective reference events, it becomes evident that coalescence (Figure \ref{fig:pdfs_merge}.e) can be predicted within a considerably narrower time window compared to nucleation (Figure \ref{fig:pdfs_nuc}.e). From this, we can conclude that the system exhibits higher variability and reduced predictability preceding nucleation events, than when predicting coalescence events.

\section{Conclusion and Future Work}

In this study, we demonstrate the ability to capture the dynamics of a stochastically driven system, namely beta-plane turbulence, using a probabilistic neural network. We demonstrate that the deep learning model remains a faithful representation of the system, even over very long time evolutions and achieves a speedup of over five-orders-of-magnitude over traditional numerical schemes. Furthermore, our work showcases the advantages of the Stochastic Latent Transformer, trained using the Continuous Ranked Probability Score and a spectral loss, when compared to existing state-of-the-art methods for probabilistic modelling of fluid flows, namely VAE and adversarial-based training. 

If the beta-plane model is accepted as a very simple analogue of mid-latitude weather variation, then the deep learning approach presented here demonstrates its capability in leaning a combined specification of large-scale dynamical evolution which includes the effects of the parameterised eddies, in contrast to work that learns just a specification of eddy forcing. This appears to provide a successful emulation of weather variability, that importantly remains faithful during long time-scales.

The speedup achieved by the neural network allows for the forecasting and characterisation of spontaneous transition events, providing probability estimates for specific lead times. This provides a tool to address questions previously unattainable with direct numerical integration methods. Our investigation highlights an asymmetry in system predictability; with the system selecting a default configuration (under the parameters used this is a 2-jet state), with nucleation events away from this state highly unpredictable, predominantly driven by stochastic forcing, while coalescence events to return to this default state act in a much more deterministic manner.

To investigate this asymmetry further, the study could be expanded across a range of parameter regimes of $\beta$ and $\mu$ to determine if the observed results persist under varying preferred jet configurations. Since the neural network is trained in a specific parameter regime, this could implement transfer learning \cite{operators_pde} to enable the model's adaptation to alternative regimes with reduced additional training requirements.

Extending this work to model beta-plane turbulence with multiple layers, to capture a system closer to observations, could be done in one of two approaches. One method involves explicit stochastic forcing to parameterise fluctuation fields, which the Stochastic Latent Transformer models, as here. In another, a system capable of generating turbulence without stochastic forcing could be employed, with the Stochastic Latent Transformer implicitly modelling the unresolved degrees of freedom to capture large-scale spatiotemporal features. The architecture is flexible to both approaches, with modifications from this work only required in the encoder and decoder to accommodate 2D and 3D inputs while retaining the same stochastic transformer in the latent space. This extension of the probabilistic deep learning model, adaptable to any SPDE, would allow for the exploration of various turbulent flows, including more complex geophysical fluid dynamics phenomena.

\section*{Open Research}

All of the code used in this paper is available under an Apache license at: \url{https://github.com/Ira-Shokar/Stochastic_Latent_Transformer}.

All of the data used in this paper is under an Apache license at:  \url{https://zenodo.org/records/10034268}.

%%%%%%%%%%%%%%%%%%%%%%%%%%%%%%%%%%%%%%%%%%%%%%%

\section*{Acknowledgements}
I.S. acknowledges funding by the UK Engineering and Physical Sciences Research Council [grant number EP/S022961/1] as part of the UKRI Centre for Doctoral Training in Application of Artificial Intelligence to the Study of Environmental Risks. We would like to thank Mat Chantry for his input and recommendations, in particular, with regard to the choice of loss function.

%% ------------------------------------------------------------------------ %%
%% References and Citations

%%%%%%%%%%%%%%%%%%%%%%%%%%%%%%%%%%%%%%%%%%%%%%%
%
% \bibliography{<name of your .bib file>} don't specify the file extension
%
% don't specify bibliographystyle

% In the References section, cite the data/software described in the Availability Statement (this includes primary and processed data used for your research). For details on data/software citation as well as examples, see the Data & Software Citation section of the Data & Software for Authors guidance
% https://www.agu.org/Publish-with-AGU/Publish/Author-Resources/Data-and-Software-for-Authors#citation

%%%%%%%%%%%%%%%%%%%%%%%%%%%%%%%%%%%%%%%%%%%%%%%

%\bibliography{enter your bibtex bibliography filename here}
\bibliographystyle{ieeetr}
\bibliography{references}

%Reference citation instructions and examples:
%
% Please use ONLY \cite and \citeA for reference citations.
% \cite for parenthetical references
% ...as shown in recent studies (Simpson et al., 2019)
% \citeA for in-text citations
% ...Simpson et al. (2019) have shown...
%
%
%...as shown by \citeA{jskilby}.
%...as shown by \citeA{lewin76}, \citeA{carson86}, \citeA{bartoldy02}, and \citeA{rinaldi03}.
%...has been shown \cite{jskilbye}.
%...has been shown \cite{lewin76,carson86,bartoldy02,rinaldi03}.
%... \cite <i.e.>[]{lewin76,carson86,bartoldy02,rinaldi03}.
%...has been shown by \cite <e.g.,>[and others]{lewin76}.
%
% apacite uses < > for prenotes and [ ] for postnotes
% DO NOT use other cite commands (e.g., \citet, \citep, \citeyear, \citealp, etc.).
% \nocite is okay to use to add references from your Supporting Information
%

\newpage

\appendix

\section{Numerical simulations of geostrophic turbulence}

Direct numerical simulation of vorticity in equation (1) is performed on a 2D doubly-periodic square domain $(x,y) \in [0, 2\pi]^2$, using a pseudo-spectral method with $N=256$ grid points - the full numerical scheme is outlined in \cite{Laura_cope}. Each of the variables can be written in terms of its discretised Fourier transform relating spatial coordinates $\textbf{x} = (x,y)$ to wavevectors $\textbf{k} = (k_x,k_y)$, together with a standard 2/3 dealiasing rule. For the case of vorticity, this is given by:

\begin{align} \tilde{\zeta}(\textbf{k}, t) = \sum_{\textbf{x} = -\pi}^{\pi} \zeta(\textbf{x}, t)e^{-i\textbf{k.x}}, \hspace{20pt} \zeta(\textbf{x}, t) = \frac{1}{N^2} \sum_{\textbf{k} = -N/2}^{N/2} \tilde{\zeta}(\textbf{k}, t)e^{i\textbf{k.x}}, \end{align}

with the allowed wavenumbers in each direction: $k_x;k_y \in \frac{1}{L_D} \left( -\frac{N}{2} +1, -\frac{N}{2} +2,..., 0, 1,...,\frac{N}{2} \right),$ where it is assumed that $N$ is even. Each vorticity equation is a stochastic partial differential equation that can be written in terms of linear and nonlinear operators:

\begin{align} \frac{\partial}{\partial t} \zeta(\textbf{x}, t) = \mathcal{L}(\zeta(\textbf{x}, t))+ \mathcal{N}(\zeta(\textbf{x}, t), \zeta(\textbf{x}, t)) + \xi(\textbf{x}, t). \end{align}

Note that the Fourier transform commutes with the linear operator but not with the nonlinear operator. Consequently, the linear and stochastic terms are time-stepped entirely in Fourier space using a fourth-order Runge-Kutta algorithm. However, the nonlinear terms must be computed in physical space before being transformed back to Fourier space:

\begin{align} \tilde{\zeta}(\textbf{k}, t+\Delta t) = \frac{\left(1 + \frac1{2} \Delta t \mathcal{L} \right) \tilde{\zeta}(\textbf{k}, t) + \left[\widetilde{\mathcal{N}(\zeta, \zeta)}(\textbf{k}, t + \frac1{2} \Delta t) + \tilde{\xi}(\textbf{x}, t) \right] \Delta t} {1 - \frac1{2}\Delta t \mathcal{L}}, \end{align}

To promote the spontaneous emergence of zonal flows, rather than being directly forced, the fluid is stirred using a stochastic vorticity force $\xi(\textbf{x},t)$ with zero mean ($\langle \xi(\textbf{x},t) \rangle = 0$) injected onto an annulus of wavevectors in Fourier space centred around a mean radial wavenumber $k_f$ with thickness $\delta k$ = 1. To prevent the direct forcing of purely zonal or purely meridional flows, we exclude wavevectors of the form $\textbf{k} = (0, k)$ and $\textbf{k} = (k, 0)$. The stochastic forcing is expressed as:

\begin{align} \xi(\textbf{x},t) = \sqrt{\frac{2\varepsilon k_f^2}{\mathcal{N}\Delta t} \left(\frac{1-\gamma}{1+\gamma} \right)} \sum_{|k - k_f|<\delta k} \tilde{\eta}(\textbf{k}, t)e^{i\textbf{k.x}}. \end{align}

where $\varepsilon$ is the energy injection rate, $\mathcal{N}$ is the number of forced wavevectors that in Fourier space, with Markovian coefficients $\tilde{\eta}$ satisfying the Hermitian property ($\tilde{\eta}(\textbf{-k},t) = \tilde{\eta}^*(\textbf{k},t)$) to ensure that the forcing function $\xi(\textbf{x},t)$ is real. Using a decorrelation time $\rightarrow 0$, the forcing takes the form of white noise with $\tilde{\eta}(\textbf{k},t) = X_n$, where the $X_n$ are independent, identically distributed random variables given by $X_n = e^{i\theta_n}$ with uniformly distributed random phase $\theta_n \in [0, 2\pi)$, and have mean $\langle X_n \rangle = 0 $ and covariance $\langle X_m X_n^* \rangle = \delta_{mn}$. 

Hyperviscosity, $\nu_n \nabla^{2n} \zeta$, is introduced to mitigate energy accumulation in small-scale fluctuations near the largest retained wavenumber, $k_{max}$, implemented as:

\begin{align} \nu_n = \frac{(-1)^{n+1} \nu}{k_{max}^{2n}} \end{align}

where $\nu$ is the hyperviscosity coefficient and $n$ is the hyperviscosity index. Simulations start from rest with $u = \zeta = 0$, and parameters remain constant throughout the integration.

The time for total kinetic energy to reach a statistically steady state during spin-up depends on the linear damping rate $\mu$, typically achieved by $\mu t = 2 - 3$ (here $t = 50 - 75$). Only data points beyond this point are used for training and testing the deep learning model.

\section{Comparisons With Other Architectures}

In Figure \ref{fig:long_emulations} we can see that the SLT produces qualitatively compelling results. However, when using other neural network architectures we would not reproduce this stability over long-time evolutions. An established method for producing probabilistic generative models is the Variational Autoencoder (VAE). Recent research \cite{vaetransformer2023}, used a combination of a VAE with a Transformer evolving the latent variables. Within this framework, the VAE maps each latent variable to a distribution parameterised by its mean, $\mu$, and standard deviation, $\sigma$, as contrasted to a single vector. Samples are subsequently extracted from this distribution, $Z = \mu + \sigma \cdot \epsilon$ with $\epsilon \sim \mathcal{N}(0, 1)$.

For comparison, we replace the Stochastic Transformer in the SLT architecture with a multi-layer LSTM that takes in $Z_{t:t-L}'$, after being phase-aligned. The LSTM outputs $\mu$ and $\sigma$, which are then sampled $\mu_{t+1} + \sigma_{t+1} \cdot \epsilon$, and passed to a nonlinear layer, followed by a linear layer, to output $Z_{t+1}'$. The phase $\phi(t)$ is reintroduced as with the SLT, before being passed to the decoder, as previously.

We found that using a VAE led to instability in autoregressive outputs as shown in Figure \ref{fig:long_emulations}.c. Upon investigation, the root cause was determined to be the prevalent blurring effect, a challenge frequently encountered with VAEs. This occurs as the VAE is trained to minimise the Evidence Lower Bound, $\text{ELBO}= \text{MSE}(U, \tilde{U}) + D_{KL}[\mathcal{N}(\mu, \sigma)|\mathcal{N}(0, 1)]$, which is a lower bound on the intractable likelihood. The MSE reconstruction loss in physical space assumes a Gaussian distribution for $U$ and prioritises values near the mean with higher probability rather than those with greater variance, leading to a conservative model. Here, at later time steps, the model is asked to make predictions using the output from a previous time step that is outside the distribution of the training dataset due to growing errors, leading to nonphysical outputs.

To overcome this, the addition of an adversarial loss term was considered. Adversarial training methods have become very popular in producing probabilistic generative models \cite{deepminddiff}, as they add an additional classifier network that must also converge during training, ensuring that the distribution of the generative model outputs matches that of the dataset \cite{timeGAN}. This then informs the generating network (in our case the VAE) whether its samples are similar to that of the training dataset. As this secondary network is also differentiable, through backpropagation the weights of the generator network are updated to produce results closer to that of the targets from the training dataset, while the classifier improves in its ability to distinguish between the two classes, leading to both networks improving until an equilibrium is reached where the generator is able to produce samples that the classifier cannot distinguish correctly - with these samples being drawn from the same distribution as the training dataset.

The classifier network, referred to as the Discriminator with weights $\chi$, is trained using the following objective function:
\begin{align} \mathcal{L}_{D} = \mathbb{E}[logD_{\chi}(U)] + \mathbb{E}[log(1 - D_{\chi}(\tilde{U})] \end{align}

where $\tilde{U}$ is the output from the VAE, outlined above. The goal of the discriminator is for the first term in equation (B1), which is the average prediction of the discriminator on real sampled data from the training dataset, to be as large as possible, corresponding to correctly identifying real data with a high probability. The second goal of the discriminator is to correctly identify generated data from the VAE as fake with high confidence, corresponding to $log D_{\chi}(\tilde{U})$ being as small as possible. To turn the expression into a maximisation problem, the term $(log(1 - D_{\chi}(\tilde{U}))$ is used in equation (B1). The real data, $U$ from the dataset is assigned a label $1$, while the fake, generated data $\tilde{U}$ is assigned a label $0$. The Discriminator outputs a value that lies between $[0, 1]$, attempting to classify the data correctly. 

The goal of the generator (here the VAE) is for $log(1 - D_{\chi}(\tilde{U}))$ to be as small as possible- corresponding to the Discriminator thinking that the generated data is real with high confidence. The following loss function is evaluated, appending an additional loss term to the ELBO loss, weighted by a parameter, $\gamma$:

\begin{align} \mathcal{L}_{\text{Adversarial-VAE}} = \text{MSE}(U, \tilde{U}) + D_{KL}[\mathcal{N}(\mu, \sigma)|\mathcal{N}(0, 1)] + \gamma \mathbb{E}_{\tilde{U}}[log(1 - D(\tilde{U})] \end{align}

However, this approach is not without its challenges, predominantly, issues of non-convergence and mode collapse may arise. The non-convergence issues arise, partly due to the fact that learning the discriminator from data with few steps of an optimisation procedure and subsequently using that to obtain the gradient of the objective with respect to the generator leads to biased gradient estimates. Mode collapse is a consequence of unstable adversarial training, in which the generative distribution collapses to a single point. In Figure \ref{fig:long_emulations}.d we observe mode collapse in the form of cyclical repeating features with a period of $~1500$ observable timesteps, where the generative model has learned dynamics that will fool the discriminator network but are physically unrealistic.

\newpage
\section{Supplementary Figures}

\begin{figure}[htbp] 
\includegraphics[width=\textwidth]{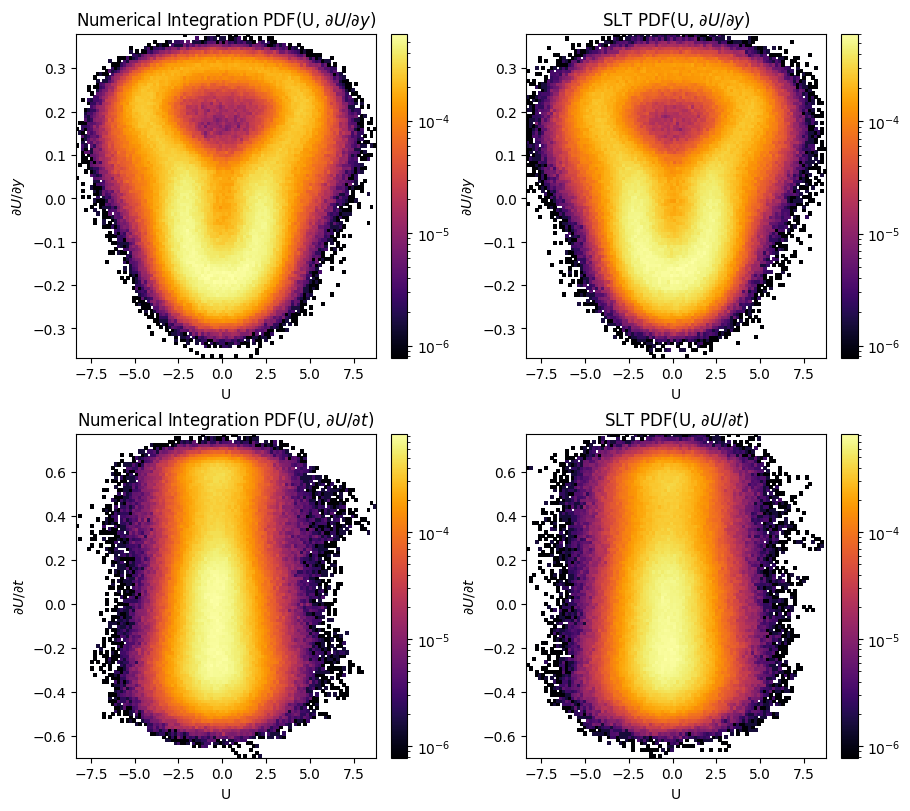}
\caption{\small \textbf{PDFs displaying joint probabilities of $U$, $\partial_y U$ and $\partial_t U$}. \textbf{a}, for visualisation of the 3D PDF used to evaluate the Hellinger distances in Figure \ref{fig:evals_long}, we average over $\partial_t U$ to obtain the density of values in the $u$-$\partial_yU$ space computed from the $\text{10}^{\text{6}}$ time steps via numerical integration. \textbf{b} shows the same $u$-$\partial_yU$ space from an evolution generated using the SLT with latent dimension size 64. \textbf{c} shows the $u$-$\partial_tU$ space by averaging over $\partial_y U$ for the numerical integration values and \textbf{d} shows the same as \textbf{c} for an evolution generated using the SLT. Here we can see that again the SLT shows good agreement with the numerical integration, covering the full space of dynamics.}
\label{fig:si_pdf}
\end{figure}

\begin{figure}[htbp] 
\includegraphics[width=\textwidth]{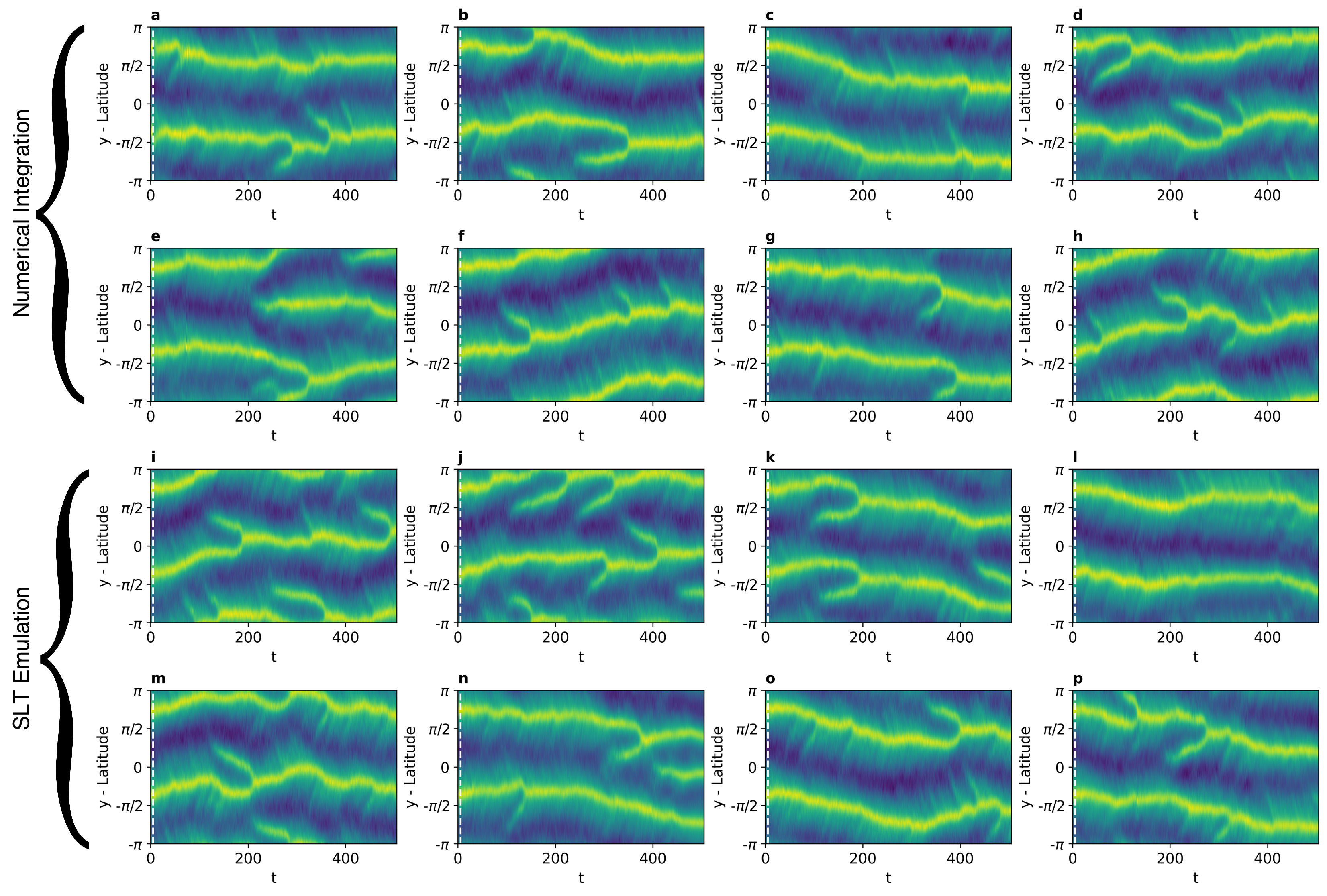}
\caption{\small\textbf{Latitude-time plots of $\mathbf{U(y,t)}$ showing a different set of initial conditions to those in Figure \ref{fig:num_int_and_ml_emul}.} \textbf{a-h} exhibit numerical integrations with identical initial conditions up to $t=10$ (indicated by the dotted line) and distinct realisations of random noise, $\xi$, after $t=10$, spanning a forecast period of 500 time units. \textbf{i-p} showcase neural network emulations with identical initial conditions as (\textbf{a-h}), but with different noise histories $\epsilon \sim \mathcal{N}(0,1)$ after $t=10$. Here the initial conditions shown to the model (left of the dashed line) are different to those Figure \ref{fig:num_int_and_ml_emul} displaying model performance in an initial two-jet state, subject to the more stochastically driven dynamics before a nucleation event.}
\label{fig:si_lat_time}
\end{figure}

\begin{figure}[htbp] 
\includegraphics[width=\textwidth]{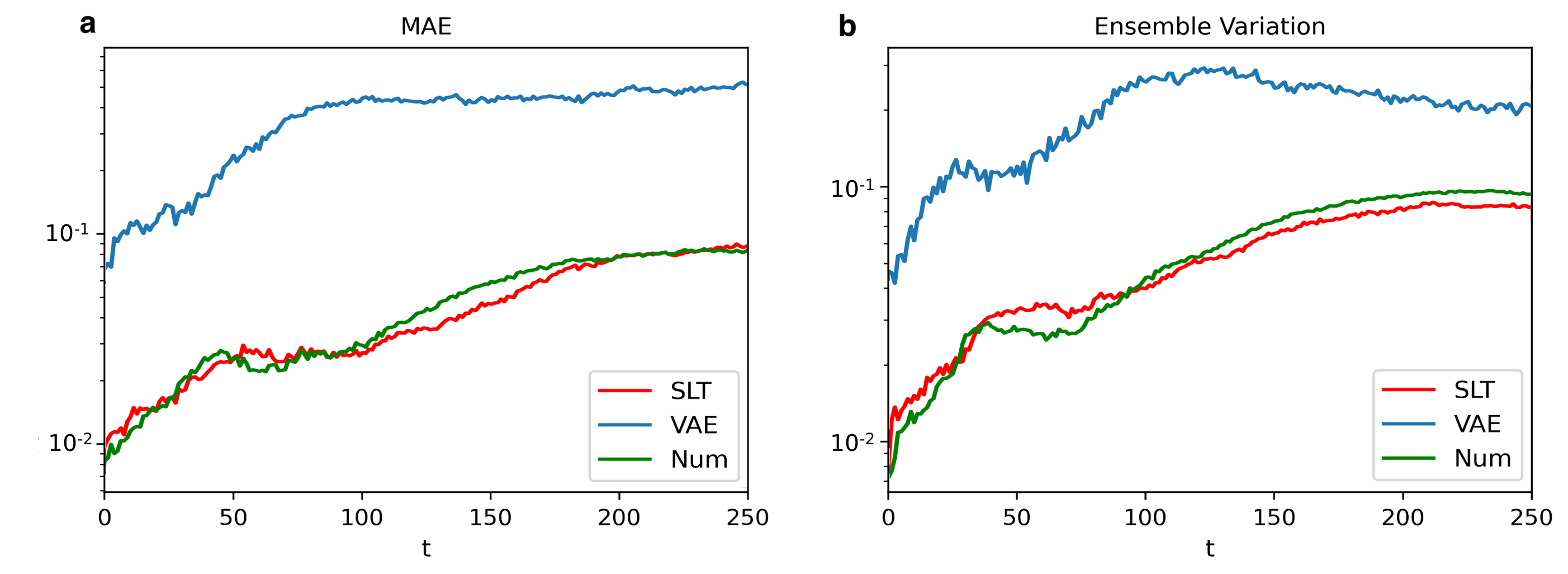}
\caption{\small\textbf{Evaluative Metrics of short-term evolutions, showing the same information as in Figure \ref{fig:evals_short}a-b, but for the ensembles shown in Figure \ref{fig:si_lat_time}. a} presents the Mean Absolute Error (MAE) for short-term tracking ability. Green represents an ensemble of 7 numerical integrations (Figure \ref{fig:si_lat_time}.b-p), while red shows 7 emulations from the SLT (Figure \ref{fig:si_lat_time}.i-o) with respect to the reference truth trajectory (Figure \ref{fig:si_lat_time}.a). Blue indicates an ensemble produced by a Variational Autoencoder (VAE). \textbf{b} showcases ensemble variation. We again show that the SLT shows very good agreement with the numerical integration.}
\label{fig:si_eval}
\end{figure}

\end{spacing}
\end{document}